\documentclass[10pt,twocolumn,letterpaper]{article}

\usepackage[pagenumbers]{iccv} 

\definecolor{iccvblue}{rgb}{0.21,0.49,0.74}
\usepackage[pagebackref,breaklinks,colorlinks,allcolors=iccvblue]{hyperref}
\usepackage{bm}
\usepackage{times}
\usepackage{epsfig}
\usepackage{graphicx}
\usepackage{bbm}
\usepackage{amsmath}
\usepackage{amssymb}
\usepackage{mathrsfs}
\usepackage{amsthm}
\usepackage{multicol}
\usepackage{multirow}
\usepackage{booktabs}
\usepackage{makecell}
\usepackage{bbding}
\usepackage{colortbl}
\usepackage{algorithm}
\usepackage{listings}

\definecolor{commentcolor}{RGB}{110,154,155}   
\lstdefinestyle{mystyle}{
    backgroundcolor=\color{white},   
    commentstyle=\color{commentcolor},
    keywordstyle=\color{purple},
    numberstyle=\tiny\color{gray},
    stringstyle=\color{blue},
    basicstyle=\ttfamily\footnotesize,
    breakatwhitespace=false,         
    breaklines=true,
    columns=fullflexible,
    escapeinside=``,
    captionpos=b,                    
    keepspaces=true,                
    numbersep=5pt,                  
    showspaces=false,                
    showstringspaces=false,
    showtabs=false,                  
    tabsize=4
}
\def\paperID{***} 
\def\confName{ICCV}
\def\confYear{2025}

\title{Training-Free Class Purification for Open-Vocabulary Semantic Segmentation}

\author{
Qi Chen\textsuperscript{1,2} \quad
Lingxiao Yang\textsuperscript{1} \quad
Yun Chen\textsuperscript{3} \quad
Nailong Zhao\textsuperscript{4} \quad
Jianhuang Lai\textsuperscript{1} \\
Jie Shao\textsuperscript{2} \qquad\qquad
Xiaohua Xie\textsuperscript{1}\thanks{Corresponding Author}\\
\normalsize\textsuperscript{1}Sun Yat-sen University \quad
\textsuperscript{2}ByteDance Intelligent Creation \quad
\textsuperscript{3}University of Surrey \quad
\textsuperscript{4}Alibaba Cloud Computing \\
\url{https://github.com/chenqi1126/FreeCP}
}

\begin{document}

\maketitle

\begin{abstract}
Fine-tuning pre-trained vision-language models has emerged as a powerful approach for enhancing open-vocabulary semantic segmentation (OVSS). 
However, the substantial computational and resource demands associated with training on large datasets have prompted interest in training-free methods for OVSS.
Existing training-free approaches primarily focus on modifying model architectures and generating prototypes to improve segmentation performance.
However, they often neglect the challenges posed by class redundancy, where multiple categories are not present in the current test image, and visual-language ambiguity, where semantic similarities among categories create confusion in class activation.
These issues can lead to suboptimal class activation maps and affinity-refined activation maps.
Motivated by these observations, we propose FreeCP, a novel training-free class purification framework designed to address these challenges. 
FreeCP focuses on purifying semantic categories and rectifying errors caused by redundancy and ambiguity.
The purified class representations are then leveraged to produce final segmentation predictions.
We conduct extensive experiments across eight benchmarks to validate FreeCP's effectiveness. 
Results demonstrate that FreeCP, as a plug-and-play module, significantly boosts segmentation performance when combined with other OVSS methods.
\end{abstract}    
\section{Introduction}
\label{sec:intro}
Segmentation has achieved remarkable success with deep learning techniques \cite{chen2017deeplab,xie2021segformer,cheng2022masked}, even in challenging semi-supervised \cite{zhao2023instance,fan2023conservative} and weakly-supervised \cite{ru2022learning,chen2022sipe} settings. 
However, traditional segmentation models are limited to segmenting a small set of predefined classes within a closed vocabulary, which is much smaller than the number of classes used by humans to describe the real world.
To address this, open-vocabulary semantic segmentation (OVSS) has been introduced to segment objects using arbitrary classes described by text.

\begin{figure}[t]
	\centering
	\includegraphics[width=0.49\textwidth]{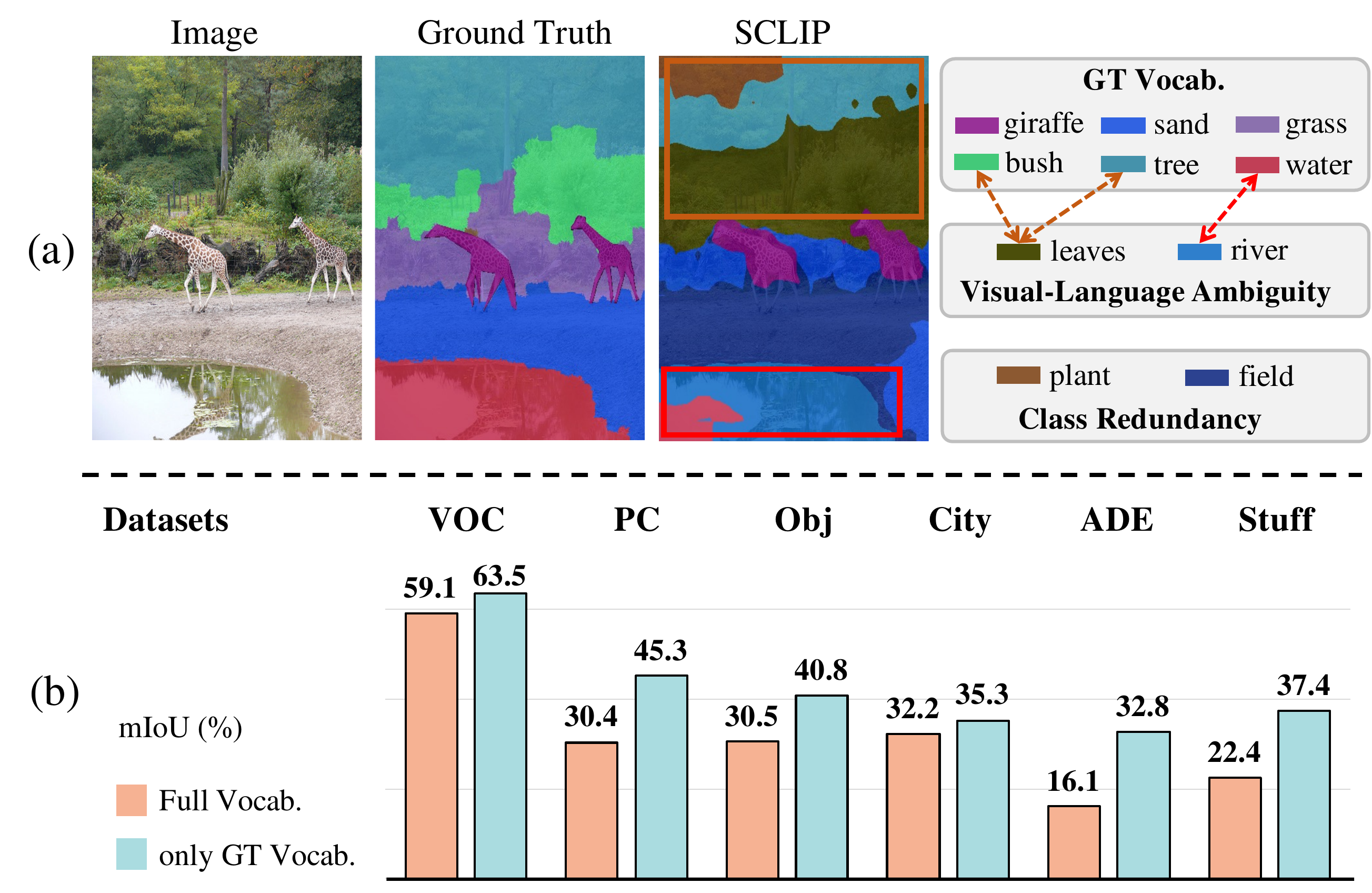}
	\caption{\textbf{Evidence} that overcomplete vocabulary affects OVSS performance. (a) Visualization of two types of problems: \textbf{Class Redundancy} and \textbf{Visual-Language Ambiguity}. (b) Performance comparison between full vocabulary and only GT vocabulary.}
	\label{fig:fig1}
\end{figure}

Large-scale vision-language models (\eg CLIP \cite{radford2021learning} and ALIGN \cite{jia2021scaling}) have demonstrated impressive transferability in recognizing novel classes, which has recently been applied in OVSS \cite{xu2022groupvit,cha2023learning,luo2023segclip}.
The mainstream approach typically involves freezing the CLIP model while training newly added modules through mask supervision \cite{ghiasi2022scaling,xu2022simple} or text supervision \cite{xu2022groupvit,liu2022open}. 
However, these methods require pixel-level annotated datasets (\eg COCO Stuff~\cite{caesar2018coco}) or large-scale image-text datasets (\eg CC-12M~\cite{changpinyo2021cc12m}), which demands significant computational resources. 
To address this limitation, recent approaches seek to exploit the localization capabilities of CLIP models to minimize training effort. 
These methods specify the forward pass for coarse localization, either by modifying the CLIP module \cite{zhou2022extract, wang2023sclip, lan2024clearclip, li2023clipsur} or integrating pretrained vision foundation models \cite{karazija2024diffusion, wang2024rim, barsellotti2024freeda, lan2024proxyclip}.

Although these methods localize objects without any training, they often yield sub-optimal results. 
The core issue lies in CLIP's design for image-text matching, which lacks the capacity for dense predictions.
Therefore, CLIP often obtains imprecise segmentation and inaccurate classification. 
As shown in \cref{fig:fig1}(a), one representative OVSS method -- SCLIP~\cite{wang2023sclip} shows the significant discrepancies between the predicted and ground-truth classes. 
These discrepancies can be categorized into two main types: \textbf{Class (Text) Redundancy}, where predictions may include classes that are not actually present in current image (\eg, `field' and `plant'); and \textbf{Visual-Language Ambiguity}, which occurs when multiple semantically similar classes are present and these classes are strongly related to the same region. 
For instance, in the \textcolor{orange}{orange} box in \cref{fig:fig1}(a), this ambiguity appears among the classes `leaves', `bush', and `tree', while in the \textcolor{red}{red} box, it occurs between the classes `river' and `water'.
To further investigate these issues, we conduct an analysis by restricting predictions to only ground-truth (GT) classes, as shown in \cref{fig:fig1}(b). 
The results reveal substantial improvements in segmentation accuracy, particularly in scenarios with large vocabulary sets, which confirms that redundant ambiguous classes significantly impair model performance. 
These findings underscore the importance of mitigating class redundancy and visual-language ambiguity to enhance segmentation performance.

\begin{figure}[t]
	\centering
	\includegraphics[width=\linewidth]{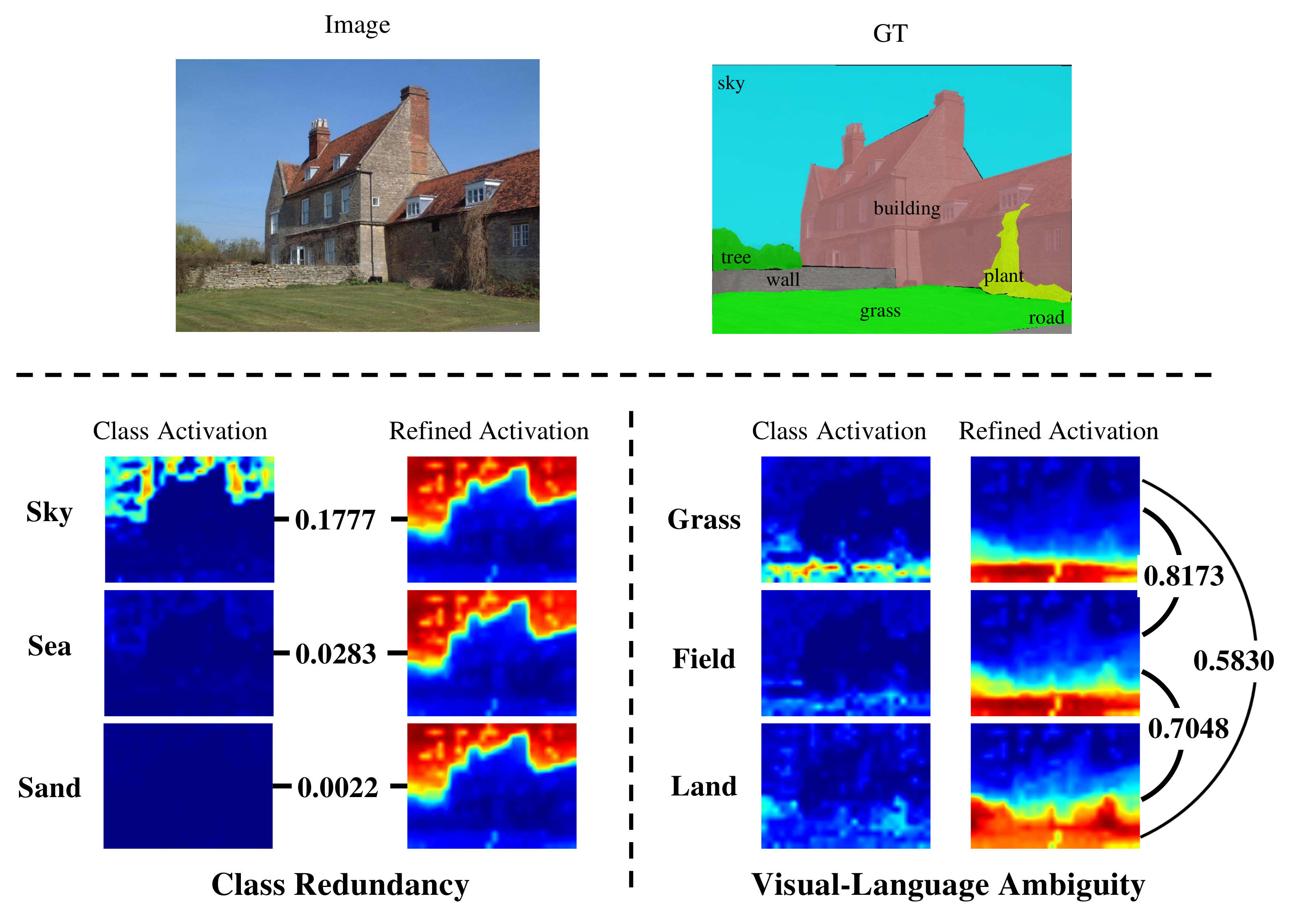}
	\caption{\textbf{Motivation of the proposed FreeCP method.} CAM and its refined version facilitate the identification of class redundancy and visual-language ambiguity by analyzing spatial distribution patterns. The reported values represent intersection-over-union (IoU), which quantify the consistency between activation maps, providing insights into the degree of similarity.}. 
    \vspace{-10pt}
	\label{fig:fig2}
\end{figure}

A naive approach is to first perform image-level multi-label classification and then use the recognized classes for segmentation. 
However, since classification occurs at the global level (image-level), it often misidentifies small or less prominent object classes as redundant and lacks the ability to detect semantically ambiguous classes in local regions.
Alternatively, bottom-up methods first segment the image and then classify each segmented region~\cite{rewatbowornwong2023zero,kang2024lavg}. 
However, the performance of these approaches is highly dependent on the initial segmentation granularity, which can lead to fragmented results.
In this paper, we aim to integrate classification and segmentation by leveraging classification cues to extract valuable local information, which is then utilized to refine class discrimination.
On one hand, Class Activation Mapping (CAM)~\cite{Zhou_2016_CVPR} can harness the CLIP model's inherent discriminative classification capabilities for dense feature localization.
Building upon this foundation, CLIP's intrinsic self-attention mechanism can further refine the CAM-generated activation patterns, enhancing the model's capacity to capture discriminative spatial-semantic features~\cite{ru2022learning,lin2023clip}.
On the other hand, the class-related spatial distribution cues provided by CAM and its refined version enable the identification of classes redundancy and visual-language ambiguity. 
As illustrated in Figure~\ref{fig:fig2}, it is achieved through quantitative analysis of spatial distribution differences between original CAM and refined CAM: 1) Class redundancy, where significant morphological variations emerge in activation regions of individual classes before and after refinement; 2) Visual-language ambiguity, manifested through substantial spatial overlap in activation patterns across multiple classes. 

Motivated by these findings, we propose a training-\textbf{Free} \textbf{C}lass \textbf{P}urification (FreeCP) method to enhance the performance of OVSS by purifying redundant and ambiguous classes.
FreeCP begins by extracting image and text token features from CLIP, and then uses these features to compute image self-affinity and image-text affinity.
The image-text affinity is then utilized to generate class-specific activation maps, which are further refined using image self-affinity.
Subsequently, redundancy purification and ambiguity purification are applied in sequence to filter out redundant classes and resolve visual-language ambiguity. 
The remaining refined activations are finally used for segmentation prediction.
Our approach achieves state-of-the-art performance on eight mainstream benchmarks, demonstrating the effectiveness of the proposed purification strategy.
The main contributions of this paper are summarized as follows:

\begin{enumerate}[leftmargin=*,label=\textbf{\textbullet}]
\item We identify the problem of the class redundant and visual-language ambiguity caused by overcompleted vocabulary in OVSS, and provide an in-depth motivation by analyzing class activation maps.
\vspace{1mm}
\item We propose a novel training-free framework, FreeCP, to purify classes. FreeCP first filters out redundant classes by examining spatial consistency between activations before and after refinement. Subsequently, it performs fine-grained recognition based on inter-class relationships to further solve visual-language ambiguous.
\vspace{1mm}
\item Extensive experiments across eight benchmarks have showcased the state-of-the-art performance achieved by our FreeCP. And the generalization ability and effectiveness of class purification have been demonstrated.
\end{enumerate} 

\section{Related Works}
\label{sec:works}
Existing OVSS methods can be categorized by their training approaches into four groups: fully-supervised, weakly-supervised, unsupervised, and training-free. 
Fully-supervised OVSS methods~\cite{li2022languagedriven,ding2022decoupling,xu2023side,xu2023open,cho2023cat,ding2023open,yu2024convolutions} initialize the model with pre-trained CLIP and then train it on large segmentation datasets. 
Weakly-supervised OVSS methods~\cite{xu2022groupvit,cha2023learning,xu2023learning,luo2023segclip,zhang2024uncovering,xing2024rewrite,ranasinghe2023perceptual,chen2023exploring,cai2023mixreorg,yi2023simple,ren2023viewco,wysoczanska2023clip} use image-text pairs as supervision, employing contrastive training to improve segmentation performance.
Unsupervised methods~\cite{zhou2022extract,chen2023exploring,wu2023clipself} use self-supervised techniques to enhance CLIP's dense prediction capabilities, avoiding the need for large image-text pair datasets.
However, these methods still require substantial training computations, resulting in significant computational overhead.

Training-free methods require no additional training and have become a popular trend in the OVSS.
MaskCLIP~\cite{zhou2022extract} modifies the self-attention layer of CLIP's vision encoder by removing the self-attention pooling layer to produce pixel-level feature maps. 
CLIP Surgery~\cite{li2023clipsur}, SCLIP~\cite{wang2023sclip}, GEM~\cite{bousselham2023grounding}, ProxyCLIP~\cite{lan2024proxyclip}, and CLIPtrase~\cite{shao2024explore} extend self-attention to more flexible and general formats, improving the segmentation ability.
CLearCLIP~\cite{lan2024clearclip} discards the residual connection and the feed-forward network of CLIP to achieve clearer and more accurate segmentation.
Another line of work uses prototypes to leverage the robust correspondence in image representations for segmenting target objects. 
ReCo~\cite{shin2022reco} employs CLIP to curate reference embeddings from unlabeled images, enhancing segmentation for rare concepts. 
Utilizing generative models, OVDiff~\cite{karazija2024diffusion} generates synthetic visual references from large text collections using diffusion models~\cite{rombach2022high}, and retrieves references based on input text for prototype-based segmentation.
Additionally, FreeDA~\cite{barsellotti2024freeda} visually localizes generated concepts, matching class-agnostic regions with semantic classes through local-global similarity.
Further advancements include RIM~\cite{wu2024image}, which integrates DINOv2~\cite{oquab2023dinov2} and SAM~\cite{kirillov2023sam} to construct well-aligned intra-modal reference features. 
CaR~\cite{sun2023car} introduces a recurrent framework that filters irrelevant text progressively.
PnP-OVSS~\cite{jiayun2023plug} combines text-to-image attention and salience dropout to iteratively acquire accurate segmentation of arbitrary classes. 
However, these methods often overlook redundancy and ambiguity among provided classes, which can lead to misclassification in localized areas. 

\section{Methods}
\label{sec:method}
In this section, we present our overall framework, FreeCP, as illustrated in~\cref{fig:fig3}. We begin by introducing fundamental formulation, CLIP and CAM in~\cref{sec:3.1}. Subsequently, we discuss the exploration of the activation refinement in~\cref{sec:3.2}. Finally, we provide a detailed explanation of our training-free class purification method in~\cref{sec:3.3}.

\subsection{Preliminary Background}
\label{sec:3.1}

\noindent \textbf{Problem Formulation.}
Given an image $\mathcal{I} \in \mathbbm{R}^{H\times W\times 3}$ and a set of semantic words $\mathcal{Z} \in \mathbbm{R}^{K}$, the objective of OVSS is to segment the pixels according to each word.
Ideally, in OVSS, precise knowledge of the parameter $K$ for each image or dataset is unavailable. 
Therefore, it is necessary to use a large $K$ to encompass a sufficiently comprehensive range of categories. 
This ensures that the segmentation achieved is highly detailed and class-aware, facilitating fine-grained understanding.
Previous approaches~\cite{ding2023open,yu2024convolutions} focus on improving models' performances by pixel-level labels or image-text pairs. 
In this paper, we propose a simple yet effective \textbf{training-free method}, which can be easily plugged into multiple methods for OVSS.

\begin{figure*}[t]
	\centering
	\includegraphics[width=0.9\linewidth]{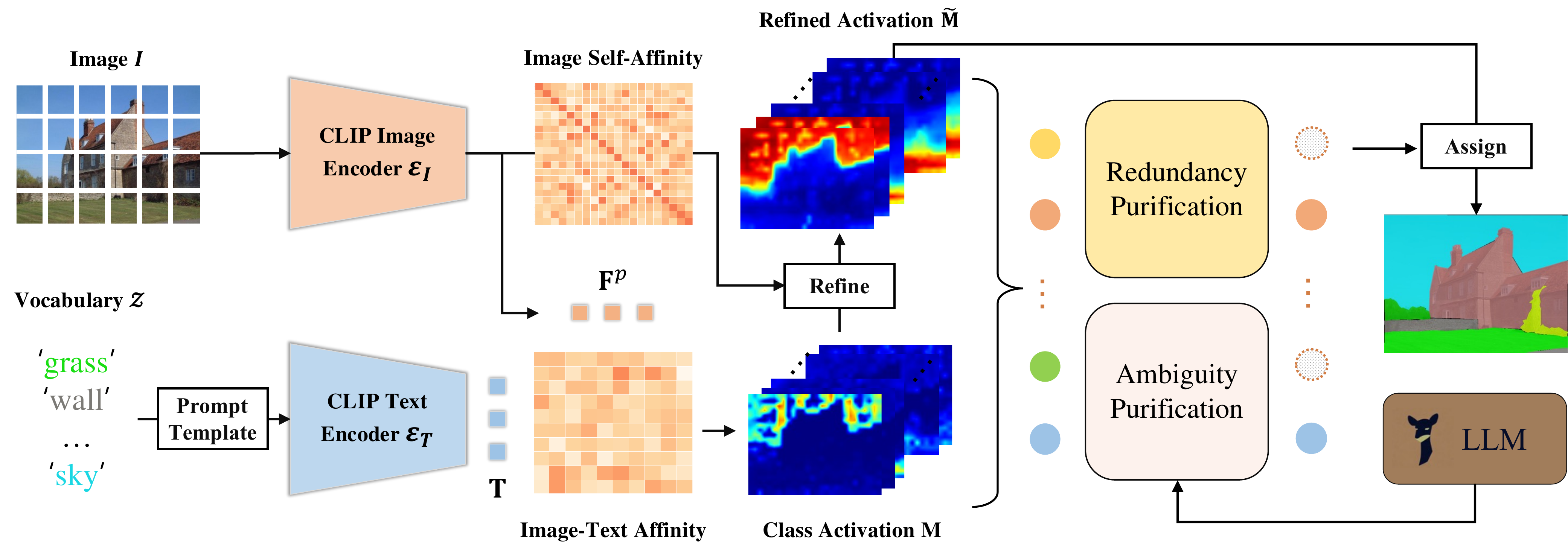}
	\caption{\textbf{Overview of the proposed FreeCP method.} 
		Based on a ViT version of CLIP model, the methodology includes three sequential stages: 
		\textbf{First}, image self-affinity and image-text affinity are derived through the image encoder $\mathcal{E}_{I}$ and text encoder $\mathcal{E}_{T}$ of CLIP. 
		We then leverage image-text affinity to generate class-specific activations $\mathbf{M}$ and refined activations $\Tilde{\mathbf{M}}$ with image self-affinity.
		\textbf{Subsequently}, we formulate spatial consistency and perform redundancy purification and ambiguity purification to eliminate redundant and visual-language ambiguous classes.
		\textbf{Finally}, the retained class activations play a pivotal role in determining the conclusive segmentation prediction.}
	\label{fig:fig3}
\end{figure*}

\vspace{1.5mm}
\noindent \textbf{CLIP-based Segmentor.}
CLIP~\cite{radford2021learning} is trained to align images and texts globally. 
Thus, it has achieved remarkable progress in image-level understanding tasks such as classification~\cite{zhou2022cocoop}, image-text matching~\cite{li2022blip}, and image generation~\cite{zhang2023formulating}.
To adapt CLIP to pixel-level prediction tasks, prior work has explored the inherent properties of the CLIP image encoder, modifying it to support segmentation. 
The mainstream approaches~\cite{li2023clipsur,bousselham2023grounding,wang2023sclip} replace $Q$-$K$ attention with $V$-$V$ attention in the last block of the self-attention module.
A few other approaches skip the last layer’s attention module entirely, taking $V$ as the output~\cite{zhou2022extract}.
Here, we adopt these modifications to enable more efficient segmentation. 
As shown in~\cref{fig:fig3}, the ViT-version of CLIP consists of an image encoder $\mathcal{E_{I}}$ and a text encoder $\mathcal{E_{T}}$. 
The text encoder $\mathcal{E_{T}}$ processes semantic words $\mathcal{Z}$ with predefined prompts, \textit{e.g.}, \texttt{A photo of a $\{\mathcal{Z}\}$}, to extract their text representations $\mathbf{T}=\mathcal{E}_{T}(\mathcal{Z}) \in \mathbbm{R}^{K\times d}$, where $d$ is the embedding dimension. 
The image encoder $\mathcal{E}_{I}$ processes $N$ patches of a single image to obtain $N$ patch tokens $\mathbf{F}^{p} \in \mathbbm{R}^{N\times d}$ and a class token $\mathbf{F}^{c} \in \mathbbm{R}^{1\times d}$.
Inspired by CAM~\cite{Zhou_2016_CVPR}, we treat the text embeddings $\mathbf{T}$ as class weights and the patch tokens $\mathbf{F}^{p}$ as image features to compute the class-wise activation map $\mathbf{M} \in \mathbbm{R}^{K\times \frac{H}{P} \times \frac{W}{P}}$, where $P$ is the patch size and $\mathbf{M}$ is defined as follow:
\begin{equation}
	\mathbf{M}_{j} = \texttt{Reshape}(\frac{\exp(\texttt{Sim}(\mathbf{F}^{p}, \mathbf{T}_{j}))}{\sum_{j}{\exp(\texttt{Sim}(\mathbf{F}^{p}, \mathbf{T}_{j}))}}),
	\label{eq:eq2}
\end{equation}
where $j \in [1, 2, \dots, K]$ is the index of the classes. $\texttt{Sim}(\cdot)$ denotes cosine similarity. $\texttt{Reshape}(\cdot)$ recovers the spatial structure of image patches by converting the 1-D patch embeddings back to 2-D maps.

\subsection{Analysis: Explore the potential of refinement}
\label{sec:3.2}

Although CLIP combined with CAM exhibits certain class-aware localization capabilities, the quality of its dense prediction remains limited due to the absence of dense supervision during CLIP's pre-training. 
Motivated by recent advances in weakly-supervised semantic segmentation~\cite{ru2022learning,lin2023clip,ahn2018learning}, which leverage affinity matrices to enhance activation maps, we investigate the efficacy of refining CAM through a learned affinity matrix. 
By propagating activations across high-affinity patches, the resulting activation maps become more comprehensive and better capture the spatial extent of object regions.
In our work, we adopt the self-attention matrices of the CLIP image encoder as the affinity matrix, capitalizing on their inherent ability to capture semantic relationships among image patches.
Specifically, the image self-affinity matrix $SA \in \mathbb{R}^{\frac{H}{16}\times\frac{W}{16} \times \frac{H}{16}\times\frac{W}{16}}$ is computed as the average of the resized self-attention matrices from multiple layers:
\begin{equation}
    SA = \frac{1}{L} \sum_{l}^{L} \psi({A_{l}}),
	\label{eq:eq2}
\end{equation}
where $A_{l}$ denotes the self-attention matrix at the $l$-th layer of the image encoder $\mathcal{E}_{I}$, $L$ is the total number of layers considered, and $\psi(\cdot)$ is a bilinear interpolation resize operation applied to standardize the spatial dimensions of attention maps across layers. 
Subsequently, the affinity-based refinement is applied to the initial activation maps $\mathbf{M} \in \mathbb{R}^{K \times \frac{H}{16} \times \frac{W}{16}}$ as follows:
\begin{equation}
	\mathbf{\Tilde{M}}_{i} = \mathbf{M}_{i}\times SA,
	\label{eq:eq3}
\end{equation}
where $i \in [1, 2, \dots, K]$ indexes the semantic classes. 

\begin{table}[t]
\centering
\caption{Comparison of using refined activation of GT vocabulary and that of full vocabulary on different datasets.}
\vspace{-5pt}
\resizebox{\linewidth}{!}{
\setlength{\tabcolsep}{1.6mm}{
\begin{tabular}{l|cccccc}
\toprule[1pt]
Methods & VOC & PC & Obj & City & ADE & Stuff \\
\midrule[1pt]
Baseline & 59.4 & 29.7 & 33.4 & 32.0 & 15.6 & 22.1 \\
Refine w/o GT & 24.0 & 19.5 & 11.3 & 23.3 & 7.6 & 13.4 \\
Refine w/ GT & 72.0 & 50.7 & 42.9 & 35.6 & 37.5 & 42.5 \\
\bottomrule[1pt]
\end{tabular}}}
\label{tab:tab1}
\vspace{-10pt}
\end{table}

\begin{figure*}[t]
    \centering
    \includegraphics[width=0.95\linewidth]{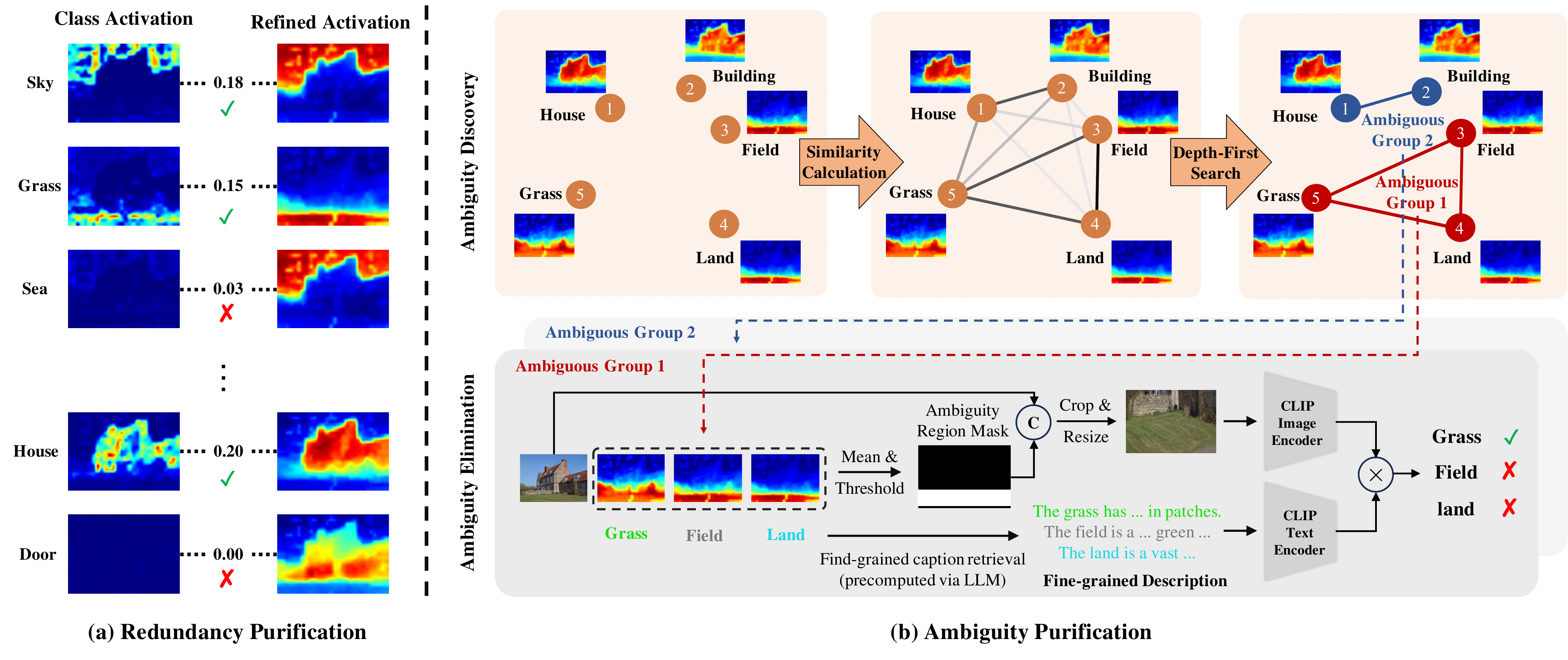}
    \caption{\textbf{Overview of Class Purification.} For all classes in the vocabulary, we first conduct (a) \textbf{Redundancy Purification}, eliminating classes whose intra-class spatial consistency falls below the predefined threshold. Subsequently, the remaining classes undergo (b) \textbf{Ambiguity Purification}, where visual-language ambiguity groups are identified based on inter-class spatial consistency. Fine-grained descriptions are then incorporated to resolve ambiguities within local regions.}
    \label{fig:fig4}
\end{figure*}

However, in open-vocabulary settings, the set of semantic categories present in an image is not known a priori, which significantly impacts the behavior of affinity-based refinement.
As shown in~\cref{tab:tab1}, when ground-truth class labels are available, the refinement process substantially improves the quality of the activation maps, as it can focus exclusively on the relevant classes, leading to more complete and accurate localization.
In the absence of ground-truth labels, the refinement results in a noticeable performance drop compared to the baseline. 
The degradation can be primarily attributed to the class-agnostic nature of the refinement mechanism, which unintentionally reinforces activation for irrelevant categories.
This issue is visually exemplified in~\cref{fig:fig2}, where refinement introduces spurious activations in regions unrelated to any ground-truth class, thereby undermining overall segmentation quality.

To address the challenges posed by affinity-based refinement in open-vocabulary settings, we analyze its behavior from the structural patterns of activation consistency across CAMs.
Specifically, we observe that:
\textit{If a class shows strong consistency in its activation map before and after refinement, it is more likely to be a true positive—that is, the class is genuinely present in the image.}
\textit{Conversely, if a class’s response changes significantly after refinement, this often indicates a redundant or irrelevant prediction.}
Furthermore, we extend this observation to inter-class relationships:
\textit{When two or more classes exhibit highly similar refined activation maps, it indicates the presence of semantic or visual-language ambiguity between them.}
These findings suggest that the consistency patterns it induces can serve as useful signals for post-hoc refinement control. 

\subsection{Training-Free Class Purification}
\label{sec:3.3}
Inspired by aforementioned observations, we propose to use spatial consistency between these activation maps to solve the class redundancy and visual-language ambiguity. 
In this paper, we introduce a simple course-level metric -- IoU to represent the Spatial Consistency (SC) between activation maps, which is defined as follows:
\begin{equation}
	\mathrm{SC}(\mathbf{X}, \mathbf{Y})= \frac{\sum[\mathbf{X}\cdot\mathbf{Y}]}{\sum[{\mathbf{X}+\mathbf{Y}} - \mathbf{X} \cdot\mathbf{Y}]}.
	\label{eq4}
\end{equation}
where $\mathbf{X}$, $\mathbf{Y}$ are the activation maps, and $(\cdot)$ represents element-wise multiplication. 

In the following, we will present our training-free two-stage class purification: \textbf{Redundancy Purification} and \textbf{Ambiguity Purification}.
Given that only a small number of classes are present in an image usually, we first identify and purify these redundant classes. Subsequently, among the remaining classes, we detect class groups exhibiting visual-language ambiguity and iteratively resolve the ambiguity using fine-grained descriptions.

\noindent \subsubsection{Redundancy Purification}
When comparing the class activation map with the refined activation map for the same class, differences in consistency appear between classes present in the image and redundant classes. 
As shown in the left side of \cref{fig:fig4}~(a), the activation maps for `sky', `grass' and `house' exhibit distinct response values, while other classes, such as `sea' and `door', display more subtle responses.
We observe that the attention-based refinement contributes to enhancing the accuracy of objects in the actual classes.
Changing before and after refinement focus on more fine details.
However, for classes like `sea' and `door', the refined activation maps introduce a significant number of irrelevant erroneous responses. 
Therefore, existing and redundant classes can be evaluated by analyzing the intra-class SC before and after refinement:
\begin{equation}
	S_{i} = \mathrm{SC}(\mathbf{M}_{i},\mathbf{\Tilde{M}}_{i}).
	\label{eq5}
\end{equation}
We empirically remove the $i$-th class if $S_{i}$ falls below a predefined threshold $T_{rp}$. 
Consequently, the updated class set \(K'\) is obtained, where \(K' \subseteq K\).

\noindent \subsubsection{Ambiguity Purification}
After filtering redundant classes, we then address the issue of visual-language ambiguity among the remaining classes.
As shown in~\cref{fig:fig4}~(b), we propose two steps to solve the ambiguity purification: discovery and elimination. 

\noindent\textbf{Ambiguity Discovery}:
To identify visual-language ambiguity, we take the inter-class spatial consistency as an ambiguity indicator.
Specifically, we calculate the SC between class $i$ and class $j$ for all possible pairs, which can be derived from \cref{eq4} and shown as follows:
\begin{equation}
	P_{i,j} = \mathrm{SC}(\mathbf{\Tilde{M}}_{i},\mathbf{\Tilde{M}}_{j}).
	\label{eq6}
\end{equation}
A higher $P_{i,j}$ indicates a greater likelihood of ambiguity between the two classes. 
Therefore, we set a threshold $T_{ap}$ to binarize this probability, highlighting potential ambiguous classes, which is defined as follows:
\begin{equation}
	P_{i,j}=\left\{
	\begin{aligned}
		1,   &\quad \quad if \quad P_{i,j} > T_{ap},\\
		0,   &\quad \quad \textrm{otherwise}.
	\end{aligned}
	\right.
	\label{eq7}
\end{equation}
Subsequently, we can extract connected class groups using a Depth-First Search algorithm based on $P$.
Each group consists of two or more mutually ambiguous classes and is defined as an ambiguity group. 
As shown in~\cref{fig:fig4}~(b), five classes are divided into two distinct ambiguity groups.

\noindent\textbf{Ambiguity Elimination}:  
For each discovered ambiguity group, we locally resolve the visual-language ambiguity with the help of fine-grained class description.
First, we need to localize the ambiguous region. 
We average activation maps of all ambiguous classes to highlight high-response regions, which likely indicate ambiguity. 
Bounding boxes of high-response regions are extracted following~\cite{lin2023clip}.
And then the ambiguous regions are cropped from the original image and resized to a specified shape (e.g., 112 $\times$ 112), forming the cropped image $\mathcal{\hat{I}}$.
With the cropped image $\mathcal{\hat{I}}$, we input it into the CLIP Image encoder to extract the visual feature of the ambiguous regions $\mathbf{\hat{F}}^{c} = \mathcal{E}_{I}(\mathcal{\hat{I}})$.

To achieve a more precise distinction among ambiguous classes, we employ a Large Language Model (LLM) to generate detailed and fine-grained text descriptions \(\mathbf{\hat{T}}_{k}\) for each ambiguous class \(k\). 
It is worth noting that for computational efficiency, we precompute and store fine-grained text descriptions for all classes within the dataset vocabulary. 
During inference, the corresponding text descriptions are retrieved directly, eliminating additional computational overhead. 
Ultimately, the final class of these ambiguous regions is determined by comparing the similarity between visual and textual features, as follows:
\begin{equation}
	{k}^{*} = \arg\max_{k}\texttt{Sim}(\mathbf{\hat{F}}^{c}, \mathbf{\hat{T}}_{k}).
	\label{eq:eq8}
\end{equation}
It is important to emphasize that no classes are removed in this step. 
Instead, classes that fail the competition are set to zero in the local region to avoid interference between verification results in different areas.

After processing all ambiguity groups, we apply the argmax operation to the refined activations of the remaining \(K'\) classes to determine the final segmentation results.

\definecolor{gray}{rgb}{0.5,0.5,0.5}
\definecolor{mygray}{gray}{0.9}
\begin{table*}[t]
\centering
\caption{\textbf{Comparison with the state-of-the-art training-free methods on eight benchmarks.} Our FreeCP can be integrated with existing training-free methods, leading to significant performance improvements on all benchmarks. Result with $^{*}$ is postprocessed with denseCRF.}
\resizebox{\linewidth}{!}{
\setlength{\tabcolsep}{2.6mm}
\begin{tabular}{lp{26pt}<{\centering}p{26pt}<{\centering}p{26pt}<{\centering}cp{26pt}<{\centering}p{26pt}<{\centering}p{26pt}<{\centering}p{26pt}<{\centering}p{26pt}<{\centering}c}
\toprule
\multirow{2}{*}{\textrm{Methods}} & \multicolumn{3}{c}{With background} && \multicolumn{5}{c}{Without background} & \multirow{2}{*}{\textrm{Avg.}}\\
\cmidrule{2-4} \cmidrule{6-10}
& VOC21 & PC60 & Object && VOC20 & City & PC59 & ADE & Stuff \\
\midrule
\multicolumn{9}{l}{\textit{With additional models, e.g., DINOv2, Stable Diffusion, SAM}} \\
ZeroGuideSeg \scriptsize{\textcolor{gray}{[ICCV'23]}}\small~\cite{rewatbowornwong2023zero} & - & - & - && 20.1 & - & 19.6 & - & - & -\\
RIM \scriptsize{\textcolor{gray}{[CVPR'24]}}\small~\cite{wang2024rim} & 77.8 & 34.3 & 44.5 && - & - & - & - & - & -\\
FreeDA \scriptsize{\textcolor{gray}{[CVPR'24]}}\small~\cite{barsellotti2024freeda} & 55.4 & 38.3 & 37.4 && 85.6 & 36.7 & 43.1 & 22.4 & 27.8 & 43.3 \\
LaVG \scriptsize{\textcolor{gray}{[ECCV'24]}}\small~\cite{kang2024lavg} & 62.1 & 31.6 & 34.2 && 82.5 & - & 34.7 & 15.8 & 23.2 & - \\
ProxyCLIP \scriptsize{\textcolor{gray}{[ECCV'24]}}\small~\cite{lan2024proxyclip} & 61.3 & 35.3 & 37.5 && 80.3 & 38.1 & 39.1 & 20.2 & 26.5 & 42.3 \\
\midrule
\multicolumn{2}{l}{\textit{Without additional models}} \\
CLIP \scriptsize{\textcolor{gray}{[ICML'21]}}\small~\cite{radford2021learning} & 18.8 & 9.9 & 8.1 && 49.4 & 6.5 & 11.1 & 3.1 & 5.7 & 14.1 \\
ReCo \scriptsize{\textcolor{gray}{[ECCV'22]}}\small~\cite{shin2022reco} & 25.1 & 19.9 & 15.7 && 57.7 & 21.6 & 22.3 & 11.2 & 14.8 & 23.5\\
CaR$^{*}$ \scriptsize{\textcolor{gray}{[CVPR'24]}}\small~\cite{sun2023car} & {67.6} & 30.5 & {36.6} && 91.4 & - & 39.5 & 17.7 & - & -\\
CLIPtrase \scriptsize{\textcolor{gray}{[ECCV'24]}}\small~\cite{shao2024explore} & 53.0 & 30.8 & 44.8 && 81.2 & - & 34.9 & 17.0 & 24.1 & - \\
OVDiff \scriptsize{\textcolor{gray}{[ECCV'24]}}\small~\cite{karazija2024diffusion} & 62.8 & 28.6 & 34.6 && 80.9 & 23.4 & 32.9 & 14.1 & 20.3 & - \\
MaskCLIP \scriptsize{\textcolor{gray}{[ECCV'22]}}\small~\cite{zhou2022extract} & 43.4 & 23.2 & 20.6 && 74.9 & 24.9 & 26.4 & 11.9 & 16.7 & 30.3\\
\rowcolor{mygray} MaskCLIP + FreeCP \scriptsize{\textcolor{gray}{[Ours]}}\small & {64.4} & {34.7}  & {36.2} && {84.1} & {32.5} & {36.6} & {17.6} & {23.3} & {41.2} \\
GEM \scriptsize{\textcolor{gray}{[CVPR'24]}}\small~\cite{bousselham2023grounding} & 56.9 & 32.6 & 31.1 && 79.9 & 30.8 & 35.9 & 15.7 & 23.7 & 38.3 \\
\rowcolor{mygray} GEM + FreeCP \scriptsize{\textcolor{gray}{[Ours]}}\small & {64.7} & {35.5}  & {36.9}  && {80.6} & {35.7} & {39.1} & {17.8} & {25.8} & {42.0} \\
ClearCLIP \scriptsize{\textcolor{gray}{[ECCV'24]}}\small~\cite{lan2024clearclip} & 51.8 & 32.6 & 33.0 && 80.9 & 30.0 & 35.9 & 16.7 & 23.9 & 38.1 \\
\rowcolor{mygray} ClearCLIP + FreeCP \scriptsize{\textcolor{gray}{[Ours]}}\small & {64.5} & {35.7}  & {36.9}  && {81.5} & {34.4} & {39.3} & {18.9} & {26.1} & {42.2} \\
SCLIP \scriptsize{\textcolor{gray}{[ECCV'24]}}\small~\cite{wang2023sclip} & 59.1 & 30.4 & 30.5 && 80.4 & 32.2 & 34.2 & 16.1 & 22.4 & 38.2 \\
\rowcolor{mygray} SCLIP + FreeCP \scriptsize{\textcolor{gray}{[Ours]}}\small & {65.8} & {35.3}  & {37.2}  && {84.3} & {33.3} & {38.0} & {18.4} & {24.9} & {42.1} \\
\bottomrule
\end{tabular}}
\vspace{-6pt}
\label{tab:tab2}
\end{table*}

\section{Experiments}
\label{sec:exp}
\subsection{Dataset and Evaluation Metric}
Similar to previous works~\cite{wang2023sclip}, we conduct experiments on five commonly used segmentation benchmarks: PASCAL VOC~\cite{everingham2015pascal}, PASCAL-Context~\cite{mottaghi2014context}, MS COCO~\cite{caesar2018coco}, ADE20K~\cite{zhou2019ade}, and Cityscapes~\cite{cordts2016cityscapes}.
\noindent\textbf{PASCAL VOC:} This object-centric semantic segmentation dataset contains 20 object classes and 1 background class. 
There are two variants of VOC: \texttt{VOC21}, which includes all 21 classes, and \texttt{VOC20}, which removes the background class to form 20 classes.
\noindent\textbf{PASCAL-Context:} This dataset contains 5,105 validation images, with 459 classes. 
The most frequent 59 classes are used to form the \texttt{PC59} version for evaluation, while we also evaluate the \texttt{PC60} by treating all other classes as background.
\noindent\textbf{MS COCO:} This dataset comprises 5,000 densely annotated validation images, including 80 thing classes and 91 stuff classes, collectively forming the \texttt{Stuff} dataset. 
\texttt{Object} merges all stuff classes into a single background class, resulting in a total of 81 classes.
\noindent\textbf{ADE20K:} This scene-parsing dataset includes 150 fine-grained classes. 
We evaluate 2,000 validation images, with an average of 9.9 classes per image.
\noindent\textbf{Cityscapes:} It is designed for urban scene understanding, including 500 validation images with 19 semantic classes.
\noindent\textbf{Evaluation metric:} For all experiments, we use mean Intersection-over-Union (mIoU) as the evaluation metric.

\subsection{Implementation details}
We adopt the ViT-B/16-based pre-trained CLIP as our default backbone. 
For generating fine-grained descriptions, we utilize the Vicuna-13b-1.5~\cite{vicuna2023}. 
We design three instructions to prompt LLM, and generate five answers for each instruction, for a total of 15 descriptions for each class. 
The final textual feature is the average of these 15 descriptions.
The prompts and samples of generated descriptions are presented in supplementary materials.
In inference, all images are resized such that the shorter side is 448 pixels (560 for Cityscapes), and a sliding window approach with a size of 384 pixels and a stride of 112 pixels is employed. 
The thresholds (\(T_{rp}\), \(T_{ap}\)) are configured
based on prior knowledge regarding the dataset’s semantic complexity and category granularity. 
For complex datasets like ADE and Stuff, which include 150+ fine-grained categories and dense scenes, a low threshold helps retain small or rare regions. 
Cityscapes and Context have moderate category counts and structured layouts, so we use a balanced threshold to maintain both precision and recall. 
Simpler datasets like VOC and Object, with fewer object-centric classes and clear foregrounds, benefit from a higher threshold to suppress noise. 
Our results do not involve any post-processing methods such as PAMR~\cite{araslanov2020single} or denseCRF. 
All evaluations are conducted using 8 × NVIDIA RTX 3090 GPUs.

\subsection{Comparison with state-of-the-art methods}
\noindent\textbf{Compared baselines.}
Following \cite{lan2024clearclip, wang2023sclip}, we compare our FreeCP with state-of-the-art OVSS methods across eight benchmarks, including three that consider background class and five that do not.
The compared methods are classified into two groups.
The first group consists of methods that rely solely on CLIP~\cite{radford2021learning}, including MaskCLIP~\cite{zhou2022extract}, ReCo~\cite{shin2022reco}, SCLIP~\cite{wang2023sclip}, CaR~\cite{sun2023car}, GEM~\cite{bousselham2023grounding}, CLIPtrase~\cite{shao2024explore}, OVDiff~\cite{karazija2024diffusion}, and ClearCLIP~\cite{lan2024clearclip}. 
The second group comprises methods that incorporate additional vision foundation models, including ZeroGuideSeg~\cite{rewatbowornwong2023zero}, RIM~\cite{wang2024rim}, FreeDA~\cite{barsellotti2024freeda}, LaVG~\cite{kang2024lavg}, and ProxyCLIP~\cite{lan2024proxyclip}. 

\noindent\textbf{Quantitative Results.}
As a strategic plug-in approach, our FreeCP can be integrated into existing training-free methods to reduce the impact of class redundancy and visual-language ambiguity on performance.
As shown in \cref{tab:tab2}, after applying FreeCP to MaskCLIP, GEM, ClearCLIP, and SCLIP, the mIoU of the original methods improved by 10.9\%, 3.7\%, 4.1\%, and 3.9\%, respectively. 
This improvement is attributed to the combined effect of the refinement techniques and class purification, which optimize object contours and enhance segmentation quality. 
It is worth noting that although MaskCLIP initially exhibited suboptimal performance, its performance improved significantly with FreeCP. 
This demonstrates that our method is robust to initial performance.
Furthermore, FreeCP achieves comparable results on several datasets compared to methods that introduce additional models. 
All these results strongly validate the generalizability and effectiveness of our FreeCP.

\begin{table}
\centering
\caption{\textbf{Ablations on class purifications.} The results of four purification options are presented. RP is Redundancy Purification, AP is Ambiguity Purification, and FreeCP uses the RP-AP option.}
\vspace{-5pt}
\resizebox{\linewidth}{!}{
\setlength{\tabcolsep}{1.8mm}
\begin{tabular}{lccccccc}
\toprule[1pt]
\textrm{Methods} & \textrm{VOC21} & \textrm{PC60} & \textrm{Object} & \textrm{City} & \textrm{ADE} & \textrm{Stuff} \\
\midrule[0.5pt]
Baseline   & 59.8  & 31.6 & 34.5 & 32.0 & 17.2 & 23.2 \\
+ Refine  & 27.5  & 21.1 & 11.9 & 26.0 & 9.1 & 14.3 \\
\midrule
\multicolumn{7}{l}{+ Purification options} \\
\ \ \ RP     & 65.8  & 35.1 & 37.2 & 33.2 & 17.8 & 24.1 \\
\ \ \ AP     & 37.7  & 26.1 & 13.6 & 24.0 & 10.8 & 15.0 \\
\ \ \ AP-RP  & 57.3  & 33.8 & 36.7 & 32.6 & 16.8 & 23.8 \\
\ \ \ RP-AP  & {65.8} & {35.3}  & {37.2}  & {33.3} & {18.4} & {24.9} \\
\bottomrule[1pt]
\end{tabular}}
\vspace{-12pt}
\label{tab:tab3}
\end{table}

\begin{figure*}[t]
    \centering
    \includegraphics[width = 0.95\linewidth]{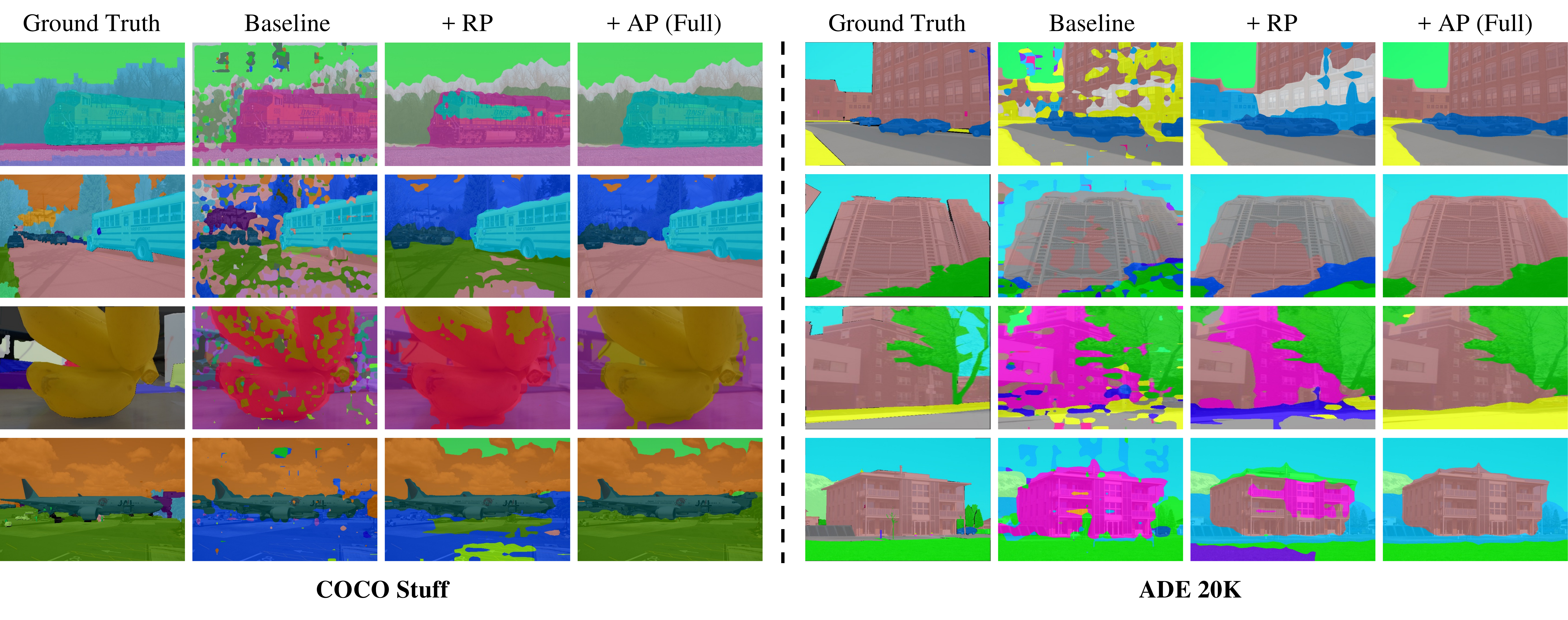}
    \caption{\textbf{Visualization of COCO Stuff and ADE20k dataset.} Our FreeCP can effectively eliminate redundancy and resolve ambiguity.}
    \label{fig:fig5}
\end{figure*}

\subsection{Ablation studies and Analyses}
In our ablation studies, we choose SCLIP~\cite{wang2023sclip} as the baseline to conduct extensive experiments, demonstrating the contribution and effectiveness of the proposed method.

\vspace{1pt}
\noindent\textbf{Effect of class purification.}
We conduct an ablation study to evaluate the effectiveness of the core contribution of our FreeCP: redundancy purification (RP) and ambiguity purification (AP). 
The comparison results are presented in \cref{tab:tab3}.
As previously discussed, refining baseline prediction without true classes leads to a sharp decline in performance.
By adding redundancy purification, we remove a significant number of weakly-responsive incorrect classes, leading to a notable improvement over the baseline.
Additionally, with the incorporation of ambiguity purification, our method achieves further improvements. 
Since class confusion is more pronounced in datasets with complex classes, the enhancement brought by ambiguity purification is relatively smaller compared to redundancy purification.
When only AP is applied, the performance significantly declines due to the persistent interference of redundant classes. 
Even if RP is performed after AP, it merely removes redundant classes without compensating for errors introduced earlier.
\cref{fig:fig5} visualizes examples from COCO Stuff and ADE20K.

\begin{table}[t]
\centering
\caption{\textbf{Ablation of textual description on ADE dataset.} We show the results of using different text for initial input and AP.}
\resizebox{\linewidth}{!}{
\setlength{\tabcolsep}{0.8mm}{
\begin{tabular}{lcc|ccccc}
\toprule[1pt]
 & Initial & AP & \textrm{{MaskCLIP}} & \textrm{SCLIP} & \textrm{GEM} & \textrm{ClearCLIP} \\
\midrule
\multirow{3}{*}{\rotatebox{90}{Baseline}} & Template & N/A & 13.0 & 17.2 & 16.6 & 17.3 \\
&Vicuna & N/A & 13.3 & 17.2 & 17.0 & 17.9 \\
&GPT-3.5 & N/A & 12.6 & 16.4 & 16.6 & 17.0 \\
\midrule
\multirow{5}{*}{\rotatebox{90}{FreeCP}} & Template & Template & 17.3 & 18.4 & 17.6 & 18.8 \\
&Template & Vicuna & \textbf{17.6} & \textbf{18.4} & 17.8 & 18.9 \\
&Vicuna & Vicuna & 17.5 & 18.2 & \textbf{18.2} & \textbf{19.4} \\
&Template & GPT-3.5 & 17.5 & 18.2 & 17.6 & 18.6 \\
&GPT-3.5 & GPT-3.5 & 17.0 & 17.8 & 17.0 & 18.4 \\
\bottomrule[1pt]
\end{tabular}}}
\label{tab:tab4}
\vspace{-10pt}
\end{table}

\vspace{1pt}
\noindent\textbf{Effect of textual description.}
\cref{tab:tab4} evaluates the impact of textual description on segmentation performance. 
In general, fine-grained descriptions consistently and significantly improve the performance of each method. 
The descriptions generated by Vicuna perform slightly better than those from GPT~\cite{achiam2023gpt}.
A key observation is that FreeCP outperforms the baseline regardless of the description used. 
This shows that the improvement is not solely due to the advantages of large language models (LLMs). 
Additionally, all methods achieve the best results when using Vicuna in the AP stage, while their preferences vary in initial textual selection.

\vspace{1pt}
\noindent\textbf{Effect of affinity features.}
We conduct an ablation study on the choice of affinity features used during refinement. As shown in~\cref{tab:tab5}, using class-agnostic masks from MaskFormer~\cite{cheng2022masked} as image self-affinity features yields suboptimal results, as these masks fail to capture contextual relationships between instances. Similarly, although DINO~\cite{caron2021emerging} and SAM~\cite{kirillov2023sam} effectively extract object-level features, they are less capable of modeling intra-class relationships.

\begin{table}[t]
    \centering
    \caption{\textbf{Ablation studies on refinement with different features.}}
    \resizebox{1\linewidth}{!}{
\setlength{\tabcolsep}{1.7mm}{
    \begin{tabular}{cccccccc}
    \toprule[1pt]
        Refine & VOC21 & PC60 & Object & City & ADE & Stuff\\
    \midrule
        MaskFormer & 43.5 & 25.2 & 23.9 & 25.8 & 12.3 & 17.3 \\
        DINO & 61.2 & 33.0 & 33.3 & 30.6 & 16.4 & 22.3 \\
        SAM & 57.3 & 31.5 & 30.8 & 29.7 & 15.2 & 21.2 \\
        CLIP & {65.8} & {35.3}  & {37.2}  & {33.3} & {18.4} & {24.9}\\     
    \bottomrule[1pt]
    \end{tabular}}}
    \label{tab:tab5}
\end{table}

\vspace{1pt}
\noindent\textbf{Effect on Different Architectures.}
Our method is compatible with various CLIP frameworks, including both ViT- and ResNet-based architectures. 
As shown in \cref{tab:tab6}, FreeCP consistently yields substantial improvements across different backbones. 
Since ResNet lacks attention layers, we employ DINO features as the affinity representation for the refinement process.

\begin{table}
\centering
\caption{\textbf{Ablation studies on different architectures.}}
\renewcommand\arraystretch{1.1}
\resizebox{\linewidth}{!}{
\setlength{\tabcolsep}{0.9mm}{
    \begin{tabular}{cc|cccccccc}
        \toprule[1pt]
          \multicolumn{2}{c}{Methods} & \textrm{VOC21} & \textrm{PC60} & \textrm{Object} & \textrm{City} & \textrm{ADE} & \textrm{Stuff}\\
        \midrule[0.5pt]
        \multirow{2}{*}{\rotatebox{0}{\footnotesize{ViT-L/14}}} & Baseline  & 46.0  & 25.8 & 25.9 & 27.9 & 15.0 & 19.9\\
        &FreeCP & 58.9  & 31.5 & 33.2 & 30.1 & 17.9 & 22.6 \\
        \midrule
        \multirow{2}{*}{\rotatebox{0}{\makecell{\footnotesize{OpenCLIP} \\ \footnotesize{ViT-L/14}}}} &Baseline  & 27.4  & 24.9 & 19.2 & 26.4 & 16.0 & 19.5\\
        &FreeCP & 57.5  & 30.7 & 32.2 & 27.7 & 18.5 & 22.5 \\
        \midrule
        \multirow{2}{*}{\rotatebox{0}{\footnotesize{R50x16}}} & Baseline & 38.4 & 18.5 & 23.7 & 17.8 & 11.8 & 15.2 \\
        &FreeCP & 52.6 & 27.7 & 30.2 & 20.5 & 13.0 & 17.6 \\     
        \bottomrule[1pt]
\end{tabular}}}
\label{tab:tab6}
\vspace{-10pt}
\end{table}

\section{Conclusions}
\label{sec:con}

In this paper, we identify the core issues as class redundancy and visual-language ambiguity, often arising from the overcomplete vocabulary. 
Based on these insights, we propose a novel training-free method, FreeCP, designed to purify classes and address these challenges. 
Extensive experiments across eight benchmarks demonstrate that our method achieves state-of-the-art performance.

\section*{Acknowledgments} 
This project is supported by the National Natural Science Foundation of China (12326618,  62206316), the Project of Guangdong Provincial Key Laboratory of Information Security Technology (2023B1212060026), the Major Key Project of PCL (PCL2024A06), and Alibaba Innovative Research Program.
Besides, the authors would like to thank Pengze Zhang (ByteDance), Shuyang Sun (Google DeepMind) and Philip Torr (University of Oxford) for their constructive assistance.

{
    \small
    \bibliographystyle{ieeenat_fullname}
    \bibliography{main}
}

\end{document}